# The Automated Mapping of Plans for Plan Recognition*

**Marcus J. Huber, Edmund H. Durfee, Michael P. Wellman**

Artificial Intelligence Laboratory
The University of Michigan
1101 Beal Avenue
Ann Arbor, Michigan 48109-2110
{ marcush, durfee, wellman } @engin.umich.edu

## Abstract

To coordinate with other agents in its environment, an agent needs models of what the other agents are trying to do. When communication is impossible or expensive, this information must be acquired indirectly via plan recognition. Typical approaches to plan recognition start with a specification of the possible plans the other agents may be following, and develop special techniques for discriminating among the possibilities. Perhaps more desirable would be a uniform procedure for mapping plans to general structures supporting inference based on uncertain and incomplete observations. In this paper, we describe a set of methods for converting plans represented in a flexible procedural language to observation models represented as probabilistic belief networks.

## 1  Introduction

Decisions about what to do should be based on knowledge of the current situation and expectations about possible future actions and events. Anticipating the actions that others might take requires models of their decision-making strategies, including models of goals that they are pursuing. Unfortunately, ascertaining the goals of others can be problematic. In competitive situations, agents may forfeit some advantage by revealing their true goals. Even in cooperative situations, explicit dialogue about goals can be impossible or undesirable given possible failures, restrictions, costs, or risks.

Agents that function in environments where explicit communication about goals is often impractical need alternative means to ascertain each others' goals, such as recognizing the plans and goals of other agents by observing their actions. To perform *plan recognition*, an observing agent needs a model of the observed agent's possible goals and plans, and what actions the observed agent could take to accomplish those plans and goals. We focus on the *case of collaborative agents*, where efficient and effective team coordination requires good models of each team participant's goals and plans. If we assume that team participants will either be designed or trained similarly, then they will have similar or identical knowledge for planning actions to achieve goals. Unfortunately, however, knowledge of the plan structures of the other agents does not, by itself permit the agent to perform plan recognition.

To perform plan recognition, therefore, an agent needs to reason from the evidence provided by observations of other agents' activities. An agent's actions are, in general, applicable toward a number of different goals, so that observation of any single action will not provide enough evidence to disambiguate the goal that motivated the agent's choice of action. Sequences of actions will tend to disambiguate the intentions of other agents, as the hypotheses that are consistent with all (or many) of the observed agents' actions gain more and more support.

An agent therefore needs to be able to take the plan structures that it has for another agent and convert them to a model that relates plans to observable actions. In this paper, we describe a method that takes plans as generated by a planning system, and creates a belief network model in support of the plan recognition task.

## 2  Related Work

An issue common to all plan recognition systems is the source and availability of the plan structure, which defines the relationships among goals, subgoals, and primitive actions. Many different plan structures have been utilized, including hierarchies of varying forms (plan spaces [CLM84], action taxonomies [KA86], AND/OR trees [Cal89], context models [Car90], plan libraries [LG91]), associative networks [CC91], SharedPlans [LGS90], plan schemas [GL90], and multi-agent

*This research was sponsored in part by NSF grant IRI-9158473, DARPA contract DAAE-07-92-C-R012, and AFOSR grant F49620-94-1-0027.

templates [AFH89]. All of these structures were designed specifically to support the plan recognition task. The direct output of a planning system, in contrast, is an object designed to be executed, not recognized. For the most part, prior work has not addressed the problem of how the plan recognition structures are (or could be) derived from executable plans as generated by planning systems.

In our research, we start from a language designed (not by us) for plan specification, as opposed to plan recognition. The particular language we have adopted is PRS [IGR92, IG90], though any standard plan language would serve just as well. PRS was chosen for a number of reasons, including that it supports all of the standard planning constructs such as conditional branching, context, iteration[1], subgoaling, etc. PRS also has a hierarchically structured plan representation which we exploit to create belief networks that are organized in a similar, hierarchical manner.

From a PRS plan, we generate a model that directly serves plan recognition by relating potential observations to the candidate plans. The model we generate is in the form of a probabilistic belief network (henceforth: belief network) [Pea88], which expresses probabilistically the causal relations among underlying goals, intentions, and the resulting observable actions.[2]

Our research bears the strongest resemblance to Goldman and Charniak's prior work on plan recognition using belief networks [CG93]. Like ours, their system generates a belief network dynamically to solve a plan recognition problem. There are several significant differences, however. First, the plan language they employ is a predicate-calculus-like representation based on collections of actions with slot fillers with hierarchical action descriptions. This representation seems well suited for modeling part-subpart relationships (goal/subgoal and is-a), and their target domain of story understanding and may have influenced this. Our plan language is based upon PRS, which has a very different set of structural primitives, including explicit sequencing, conditionalization, iteration and context. PRS is a general purpose planner, with a representation that is intended to permit any form of plan structure.

Second, Goldman and Charniak first translate plan knowledge into an associative network (their term) by using a set of generic rules for instantiating (unifying) the network with the plan knowledge. It is these instantiated rules from which they dynamically generate a belief network for a given sequence of observations

(i.e. bottom-up). Our system, on the other hand, generates a belief network from the plan representation itself, and before receiving any observations (i.e. top-down). We foresee the top-down approach having the characteristic of being able to prune (perhaps significant) portions of the resulting belief network based upon the context in which the plan recognition system finds itself. We believe these approaches are complementary, both in addressing separate sets of plan-language issues, and in emphasizing different forms of dynamism in model generation.

Finally, this work is related to a growing body of other work in the dynamic generation of belief networks [Bre92, WBG92]. Although our methods are specifically geared to plan recognition (like Goldman and Charniak's), techniques for generating probabilistic models from other forms of knowledge may have wider applicability.

## 3   PRS and Belief Networks

The Procedural Reasoning System (PRS) [IGR92, IG90] specifies plans as collections of actions organized into *Knowledge Areas*, or KAs. PRS KAs specify how plans are selected given the current goal (its *purpose*) and situation (its *context*). PRS KAs also specify a procedure, called the KA body, which it follows while attempting to accomplish its intended goal. This procedure is represented as a directed graph in which nodes represent states in the world and arcs represent actions or subgoals. Actions may consist of primitive operations (indicated by * in KA diagrams), goals to achieve (!), goals to maintain (#), goals to be tested (?), or conditions to be waited upon (ˆ). KA actions may also assert facts (→), or retract them (←). Branches in the graph may be of type AND or OR, indicating, respectively, that all or only one of the branches must be completed successfully in order to satisfy the KA's purpose. See the PRS papers [IGR92, IG90] for a more detailed description.

A belief network is a directed acyclic graph $(F, X)$ representing the dependencies $F$ among a set of random variables $X$. Each random variable $x_i \in X$ ranges over a domain of outcomes $\Omega_i$, with a conditional probability distribution $\Pi_i$ specifying the probabilities for $x_i = \omega_i$ for all $\omega_i \in \Omega_i$, given all combinations of outcome values for the predecessors of $x_i$ in the network. For a more thorough account of belief networks, see, for example, [Pea88] or [Nea90]. To avoid confusion, we refer to the action and goal nodes in a KA as *nodes*, and the nodes of a belief network as (random) *variables*.

## 4   The Mapping Method

We now describe our method for mapping plans into belief networks, first with simple sequences of actions and then with more complex plan structures. The re-sulting procedures a broad class of plans, including those with conditional branching and subgoaling. Two notable features that we do not cover, however, are iteration (or recursion), and plan variables. Both are left for future work.

---

[1]Our methodology does not currently support iteration, although this is being investigated.

[2]The issue of probabilistic plan recognition is orthogonal to the issue of probabilistic planning (cf. BURIDAN [KHW93], for example) and hence the representations created for planning under uncertainty are not inherently any more conducive to the plan recognition process.

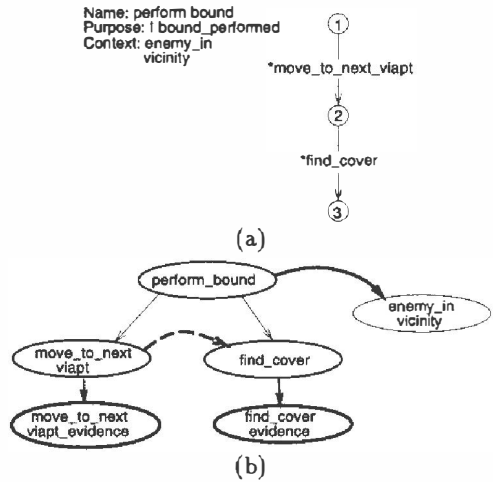

Figure 1: (a) Single level, non-branching KA. (b) Belief network.

In the remainder of the section, we discuss the basic operations involved in mapping PRS KAs to belief networks. Our description is illustrated with an example military reconaissance task, in which two (or more) cooperative agents pass through a sequence of locations, alternately navigating (also called *bounding*) or protectively watching (*overwatching*) while concealed from view.

## 4.1 Single, non-branching plans

Figure 1(a) depicts an example PRS plan consisting of a simple sequence of primitive actions. This KA says that in order to achieve the goal of accomplishing a "bound" goal, the operations of moving to the next location (the *via point*) and finding a place of concealment must be accomplished. Knowing this, if an observer were to see an agent moving toward a grove of trees, the observer might predict that the observed agent was about to enter the grove. We would like the belief network generated from this KA to support this sort of inference.

The first step in creating the belief network is to create a variable representing the goal to be achieved by the KA. The remaining variables, connections, and probabilities all provide evidence for or against the proposition that this is the goal being pursued by the observed agent. In our figures, we use the KA's name for the variable in the belief network representing the KA's goal.

We now create a new random variable for each action in the KA. The state space for each variable is determined by whether the action is a goal—with a state space of {Inactive, Active, Achieved}, or a primitive action (a basic, non-decomposable behavior of the agent)—with a state space of {Performed, NotPerformed}. Each of these new variables is dependent upon the KA's goal variable because it is the adoption of this goal that causes the performance of these actions in this particular sequence.[3] To model the temporal relationship between *move_to_viapt* and *find_cover*, we create an arc between these variables.[4]

Because we are constructing the belief network in order to perform plan recognition, it is important to model the uncertainty associated with observations [HD93]. For example, detecting the exact movements of another agent might be error-prone, while it might be easy to ascertain when the agent enters a grove of trees. Yet whether this entry represents a concealment action may be relatively less certain. To capture these differences, we add *evidence* variables to represent the relation between an observation and our belief that the observed event is an instance of the corresponding action. Evidence variables also provide a way to account for features that, while not corresponding to actions directly, provide some information regarding whether the action was performed. This indirect evidence is often all we have, as some fundamental actions may be inherently unobservable. In Figure 1(b), we indicate evidence variables by drawing them with heavy outlines.[5]

A typical KA also specifies the *context* in which it is useful, which restricts its applicability for the associated goal. For example, the "bounding overwatch" technique of travel between locations might only be necessary when enemy forces are in the vicinity. To capture these constraints in the belief network, we add one new variable for each condition in the KA's context, and include a dependency link from the goal to each context variable. The belief network constructed for the KA shown in Figure 1(a) is shown in Figure 1(b).

The last task is to determine the probability distributions for each of the random variables. Unfortunately, information about the degree of uncertainty in these relationships is not inherent in the executable plan description, and no planning system provides this probabilistic knowledge as a matter of course. We could specify this information separately based on our own subjective assessment of the domain, or it could be estimated syntactically by analyzing the frequency of occurrence of particular actions among all those that achieve particular goals. Alternately, the probabilities might be determined through empirical study of the frequency of occurrence of goals and actions during the execution of the plans in actual situations. If there is no available probabilistic information, a default assignment of equiprobability among alternatives can be used to indicate this lack of knowledge. This would permit a belief network to be fully specified in the presence of incomplete modeling information while perhaps still providing useful inferences based upon the part of the model that was specified.

---

[3]In our depiction of belief networks, we distinguish among the various sources of dependency graphically by line type: subgoal/subaction arcs are normal-weight solid lines, inhibitory arcs are normal-weight dashed lines, temporal dependency arcs are heavy dashed lines, and context arcs are heavy solid lines.

[4]To apply this technique for a plan language supporting partially ordered actions, we would simply omit the temporal dependency arcs between steps in plans that are unordered.

[5]In subsequent figures, for simplicity, we treat evidence implicitly by depicting the pair of action and evidence as a single variable.

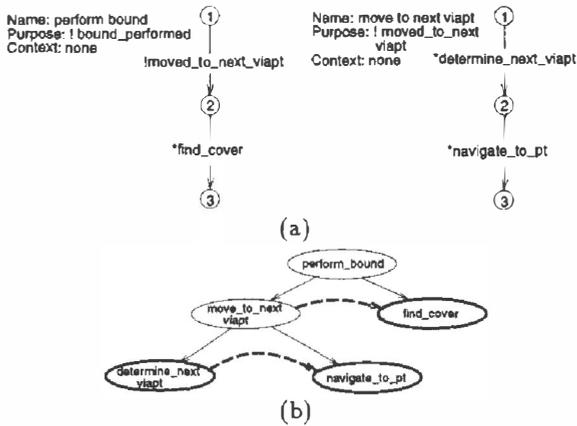

Figure 2: (a) Multi-level KA. (b) Corresponding belief network.

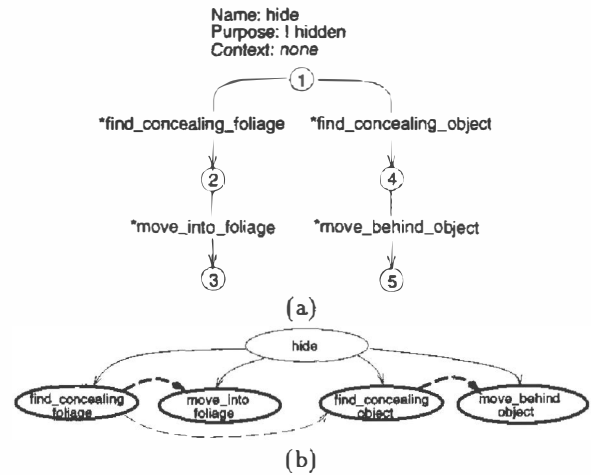

Figure 3: (a) Single plan with OR branch. (b) Corresponding belief network.

Some of the dependencies of the constructed belief network are generically specifiable, however. For example, the relation between goal and context variables (if they represent true constraints) are partially deterministic, as the goal cannot be active unless the context condition is satisfied.

The procedure for subgoaling plans is essentially the same as that for the single-level case, with the extension that subgoals need to be expanded into their constituent KA procedure. This requires treating the subgoal as a goal variable in Section 4.1. An example multi-level KA is shown in Figure 2(a), and Figure 2(b) depicts its corresponding belief network. Notice that the belief network structure beneath the *move_to_next_viapt* variable has the same form as that of *perform_bound* in Figure 1(b).

### 4.2  Conditional plans

For plans with conditional branches, the KA's goal is again the root variable for the belief network. Each action in the KA body becomes a random variable in the network as in the mapping specified in Section 4.1. However, in the conditional case, not all actions are linked. For instance, an OR branching in a KA means that an agent need only successfully execute one of those branches. We assume that one branch is executed (either successfully or unsuccess-fully) before another one is tried, so that only one sequence of actions will be active at one time. Therefore, the action variables within a branch are linked temporally as in a non-branching plan, and the variables representing the first actions in each of the disjunctive branches are linked with inhibitory arcs representing their exclusivity. The effect of this arc is that positive belief that the agent is pursuing one of the branches will inhibit belief in the alternative branch(es).[6] For AND branches, we can similarly assume either independence (our default), or a positive mutual reinforcement among branches. An example of a KA with an OR branch, and the resulting belief network, are shown in Figure 3(a) and Figure 3(b), respectively. If the branch were an AND instead, the same belief network would result, minus the arc between *find_concealing_foliage* and *find_concealing_object*.

### 4.3  Multiple goals, multiple plans

Quite often, there are several top-level goals that an agent may be pursuing. To represent the interdependencies between multiple top-level goals, we adopt the convention of always creating an arc between the top-level goal variables and modeling the dependence (or independence) through the conditional probabilities associated with these variables. An example of a mapping for this type of plan to a belief network is shown in Figures 4 and 5.

Thus far we have assumed that an agent being observed is pursuing only a single approach (KA) to satisfy each of its goals. However, there are often multiple KAs for any goal. The approach that we take is similar to the mapping for OR branches. We first create an abstract goal variable that encompasses the KAs with a common purpose (goal). The variables that represent each of the alternate plans (KAs) are then connected, with the alternate plans as the dependents, in keeping with the future expected use of the belief network. An example of multiple goals is presented in Figures 6 and 7.

---

[6] The assumption of exclusivity can be relaxed by suitable assignments of inhibitory probabilities. Or, we could alternately have chosen to assume that pursuit of the alternate branches are independent, in which case the inhibitory dependencies would be unnecessary.

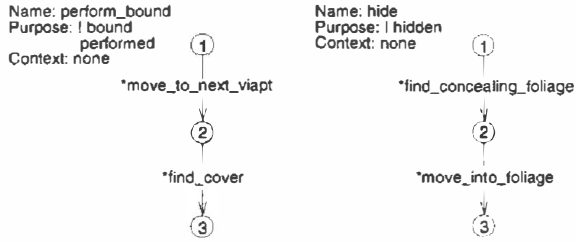

Figure 4: Multiple top-level plans.

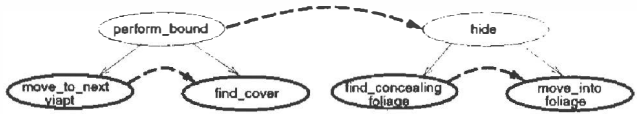

Figure 5: Belief networks for multiple top-level plans.

### 4.4   Summary

The following table (Table 1) shows a summary of the mapping methods, with the various plan features in the left column and their corresponding belief network topology in the right column.

| Plan Construct | Belief Net Topology |
|---|---|
| Subgoal/Action | new variable for subgoal/action, subgoal/action variable is child of the supergoal variable. |
| Action Sequence | new variable for each action, each action variable is child of KA's goal variable, temporal arcs between steps. |
| Context | new variable for context, context variable becomes child of goal variable. |
| OR Branch | separate action sequences for each branch, branch node variable is parent to all initial action variables of each branch, inhibitory arcs between initial action variables of each branch |
| AND Branch | same as OR branch but without the inhibitory arcs. |
| Multiple Goals | separate variable for each goal, inhibitory arcs between competing goals. |

Table 1: Mapping methodology summary.

## 5   An Example

The following example illustrates the entire process, mapping PRS plan structures to a belief network, and using the result for plan recognition.

### 5.1   Mapping to belief network

Figure 8 depicts four KAs relevant to the bounding overwatch task. The *!bound_performed* KA shows that the agent must first move to its next via point before looking for a suitable place from which to watch over the other agent. There are two KAs for dealing with an enemy agent, both conditioned on the context of an enemy agent having been sighted. Hiding, however, can consist of either moving into foliage or moving behind some concealing object. Furthermore, moving to a via point requires the agent to first accomplish *!moved_to_next_viapt*, the rightmost KA in Figure 8, which consists of a simple, non-branching sequence of operations.

Using the methods described in Section 4, the system begins mapping this collection of KAs into a belief network, starting with the top-level goals of *!bound_performed* and *!dealt_with_enemy*. The system finds that the first action in the *!bound_performed* KA is the goal *!moved_to_next_viapt* and recurses. The *!moved_to_next_viapt* KA is straightforwardly added and the mapping of *!bound_performed* resumes. The system then proceeds to map *!dealt_with_enemy*. As *!dealt_with_enemy* has two potentially applicable KAs, the methodology of Section 4.3 is used, where each KA is processed individually and then joined by an abstract goal variable representing both KAs. In addition, the OR branch in the *hide* KA complicates the construction a bit by introducing additional dependencies (as discussed above in Section 4.2). To complete the mapping, the system creates an inhibitory link between the two top-level goals (*!bound_performed* and *!dealt_with_enemy*) to indicate that only one OR the other of these goals can be achieved at the same time. The finished belief network structure is shown in Figure 9. The marginal and conditional probabilities are then loaded into the network (as mentioned in Section 4). We now show how the new representation permits an agent to infer the plans and goals of another agent based on its observed behavior.

### 5.2   Plan recognition

Suppose that Agent A is watching Agent B as they perform a reconnaisance task. Agent A and Agent B are in the military so of course there are standard operating procedures for everything. In this case the agents are using bounding-overwatch for reconnaisance, which means that one agent moves while the other agent watches for danger while concealed, with the two agents alternating between roles. These procedures are represented by the KAs in Figure 8, which get mapped into the belief network structure shown in Figure 9. Agent A observes the actions taken by Agent B, and, whenever an action is performed that fits into the model, Agent A adds this information to the belief network. After the evidence is propagated, the resulting probability distributions represent Agent A's beliefs about Agent B's goals posterior to the evidence.

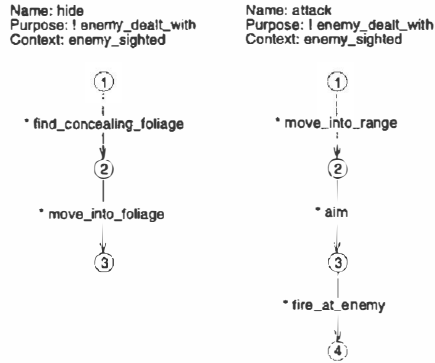

Figure 6: Multiple top-level KAs of observed agent.

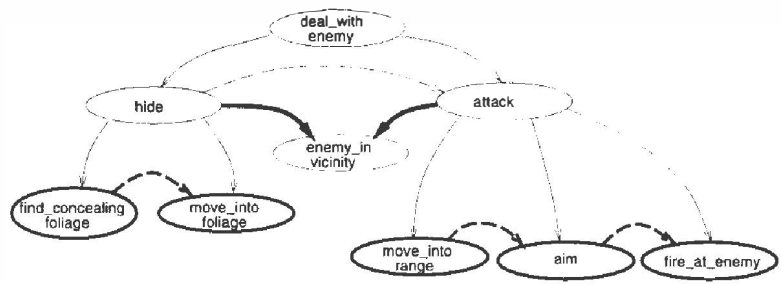

Figure 7: Belief network for multiple top-level goals.

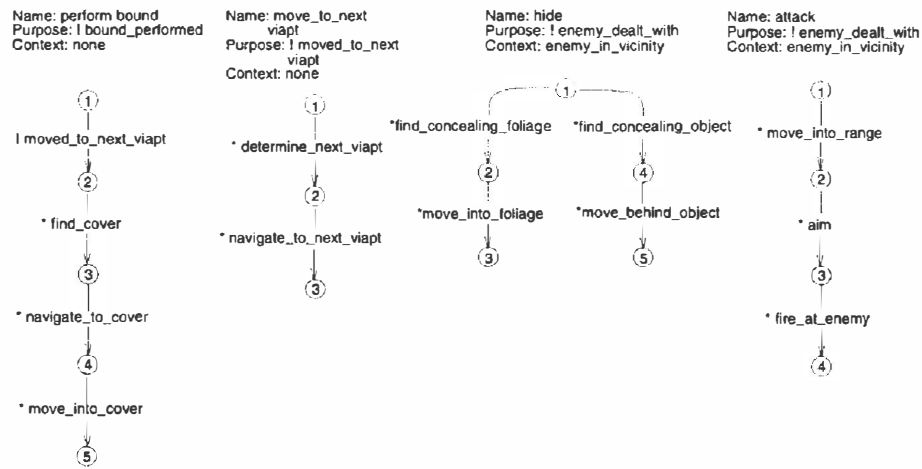

Figure 8: KAs of observed agent.

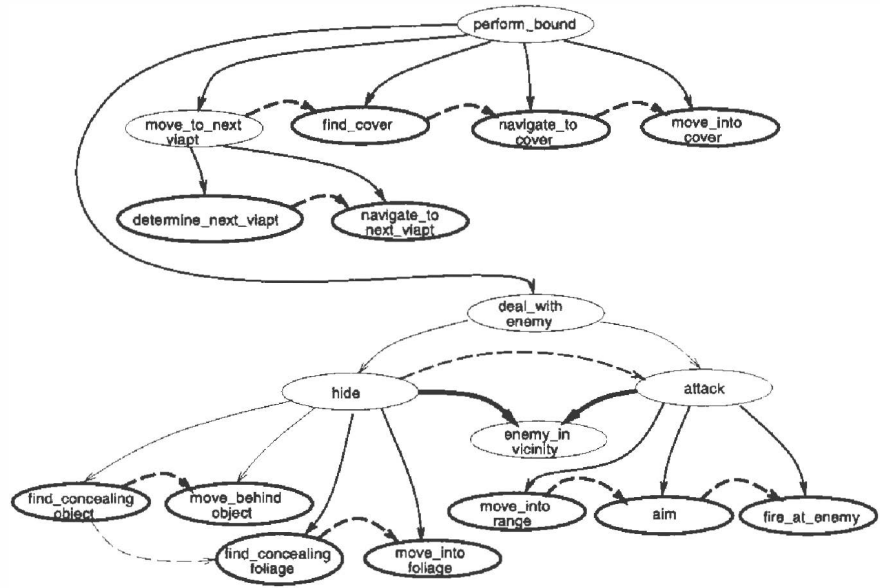

Figure 9: Final belief network representation.

Starting with a belief network without any applied evidence, suppose Agent A notices that Agent B moves in the direction of, and arrives at, the next stop in the reconnaisance route.[7] Agent A interprets this as having completed *!moved_to_next_viapt* (an example of the ability to make observations at a higher level than the "primitive" action level) and propagates this evidence through the network. This results in distributions of *perform_bound*: Inactive 0.19; Active 0.69, Achieved 0.11, and *deal_with_enemy*: Inactive 0.41; Active 0.49; Achieved 0.09. The relatively high level of belief in the *deal_with_enemy* goal is due to its having a high prior. This evidence alone suggests to Agent A that Agent B might be involved with performing the bounding role of the reconnaisance task, but it is not entirely clear at this point. Agent A determines, based upon its beliefs, that it should continue to watch for enemy agents. Continuing, if Agent A later observes Agent B moving into cover, Agent A now strongly believes that Agent B is finishing its bound process with beliefs of *perform_bound*: Inactive 0.0; Active 0.17, Achieved 0.83, and *deal_with_enemy*: Inactive 0.62; Active 0.32; Achieved 0.06. However, if instead of moving to a via point, Agent B moves in some other direction and moves into a growth of foliage, Agent A, through the plan recognition system, realizes that Agent B established a goal of *hide* (Inactive 0.03, Active 0.51, Achieved 0.46) since it has detected an enemy (Performed 0.64, NotPerformed 0.36) and that it should therefore come to its aid.

## 6    Conclusions

We have described methods by which plans in their executable form can be automatically mapped to belief networks. The examples of the implemented system illustrate that, at least for the simple plans so far explored, our methods yield belief networks that allow agents to recognize the plans of others. In the near future we plan to extend our methodology to deal with iteration and recursion, and to implement this system on physically embodied agents (robots) that will use plan recognition as part of their coordination mechanism.

While much work yet remains, we see these methods as important steps toward knowledge re-use, where automating the mapping process allows the same knowledge to be used for both planning and plan recognition. Moreover, just as concerns about storing all possible plans for all possible combinations of goals and worlds led to algorithms for dynamically constructing plans on the fly, so too do concerns about building unwieldy models of agents' actions in all possible worlds lead to a desire for dynamically constructing belief network models for situation-specific plan recognition activities. Our methods represent some initial steps in this direction.

---

[7]Until Agent B actually arrives at the via point, its movements might be ambiguous enough that it is unclear which of the move-type observations should be instantiated. In this case, evidence for *all* of them might be instantiated and the resulting beliefs used, providing Agent A with at least some information.